\documentclass{article}
\usepackage{spconf,amsmath,graphicx,xcolor}
\usepackage{booktabs} 
\usepackage{threeparttable}
\usepackage{multirow}
\usepackage{enumitem}
\setlist{nosep, leftmargin=14pt}

\usepackage{mwe} 

\title{Sparse Anatomical Prompt Semi-Supervised Learning with Masked Image Modeling for CBCT Tooth Segmentation}
\name{Pengyu Dai $^a$, Yafei Ou $^b$, Yuqiao Yang $^b$, Yang Liu$^c$, Yue Zhao $^{a, d}$
\thanks{\textit{(Corresponding authors: Yue Zhao)} (e-mail: zhaoyue@cqupt.edu.cn).} 
}
\address{$^a$ School of Communication and Information Engineering, \\Chongqing University of Posts and Telecommunications, Chongqing, 400065, China\\
$^b$ Institute of Innovative Research, Tokyo Institute of Technology, Kanagawa, 226-8503, Japan,\\
$^c$ 
Stomatological Hospital of Chongqing Medical University, Chongqing, 401147, China\\
$^d$ School of Mechanical Engineering, Zhejiang University, Zhejiang, 310058, China
}
\begin{document}
%
\maketitle
\begin{abstract}
Accurate tooth identification and segmentation in Cone Beam Computed Tomography (CBCT) dental images can significantly enhance the efficiency and precision of manual diagnoses performed by dentists. However, existing segmentation methods are mainly developed based on large data volumes training, on which their annotations are extremely time-consuming. Meanwhile, the teeth of each class in CBCT dental images being closely positioned, coupled with subtle inter-class differences, gives rise to the challenge of indistinct boundaries when training model with limited data. To address these challenges, this study aims to propose a tasked-oriented Masked Auto-Encoder paradigm to effectively utilize large amounts of unlabeled data to achieve accurate tooth segmentation with limited labeled data. Specifically, we first construct a self-supervised pre-training framework of masked auto encoder to efficiently utilize unlabeled data to enhance the network performance. Subsequently, we introduce a sparse masked prompt mechanism based on graph attention to incorporate boundary information of the teeth, aiding the network in learning the anatomical structural features of teeth. To the best of our knowledge, we are pioneering the integration of the mask pre-training paradigm into the CBCT tooth segmentation task.  Extensive experiments demonstrate both the feasibility of our proposed method and the potential of the boundary prompt mechanism.
\end{abstract}
\begin{keywords}
CBCT tooth segmentation, Self-supervised learning, Masked image modeling, Semantic segmentation
\end{keywords}

\begin{figure*}[!t]
\centering
\includegraphics[width=0.9\textwidth]{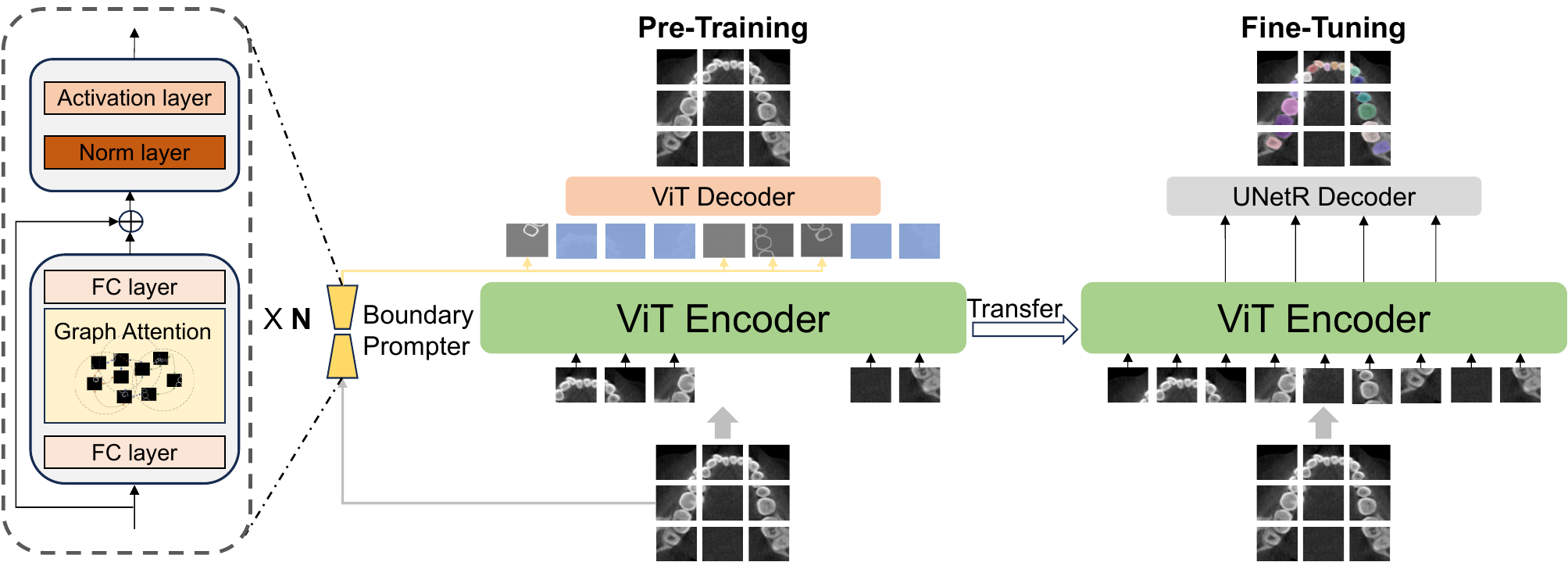}
\caption{The architecture of the proposed method. The proposed method includes three stages. First, we fix the parameters of a Graph-based boundary prompt branch, which is trained on sparse boundary annotations. Second, we self-supervised pre-train a U-Net-based masked autoencoder in an unlabeled manner. Particularly, the fixed boundary prompt branch will provide effective sparse boundary representations to the masked parts of the network in the process. Finally, we apply the pre-trained network to the downstream task of CBCT tooth segmentation.}
\label{fig_frameworks}
\end{figure*}
\section{Introduction}
\label{sec:intro}

The implementation of computer-aid digital dentistry significantly enhanced the effectiveness of diagnosis and treatment planning conducted by dental providers. Dental imaging, recognized as an indispensable diagnostic tool, has played a pivotal role in this process. In particular, Cone Beam Computed Tomography (CBCT) image, a widely used imaging modality, provides a comprehensive repository of semantic data, rendering intricate three-dimensional view of the oral cavity~\cite{9789163}. The precise segmentation of CBCT images establishes a robust foundation for a wide array of medical procedures. 
However, challenges still persist, primarily due to: 1) The time-consuming nature of semantic-level CBCT annotations, requiring specialized dental expertise.
2) Due to the close proximity of the tooth boundaries, the semantic features of them are very similar, leading to direct segmentation prone to category confusion.

Related works include Mask-RCNN-based multi-branch instance segmentation~\cite{cui2019toothnet, dentalnet}, which effectively utilizes the characteristics of the two-stage segmentation network to deeply supervise the morphological features of the segmentation results by adding additional branches. UNet-based~\cite{3dunet} semantic segmentation strategies~\cite{vnet,li2022semantic}, which facilitate network convergence through multi-stage incremental training or introduction of traditional image algorithms. Multi-network, multi-stage collaborative strategy~\cite{cui2022fully,jang2021fully}, which cascade through a multi-stage backbone network for fine-grained division of the segmentation task as a solution to realize high-precision tooth segmentation. Nevertheless, while the aforementioned methods successfully achieve precise tooth segmentation from various angles, they still demand a substantial volume of finely annotated data and struggle to effectively harness the substantial amount of unlabeled data available.

Self-supervised learning~\cite{zhai2019s4l} presents a viable alternative, as it derives its supervisory signals directly from the unlabeled data. It has recently demonstrated its effectiveness in mitigating the need for vast data quantities and in acquiring transferrable, high-dimensional representations of input data~\cite{madan2023self}. Among these approaches, one noteworthy learning task is masked image modeling~\cite{he2022masked}, wherein a subset of input patches is concealed, and the model aims to predict these masked patches. Notably, this paradigm has yielded remarkable success. Such methods have shown potential in medical images~\cite{zhou2022self,MIM}, however, distinguished from soft tissue modality medical images, CBCT oral images have non-uniform intensity variations, resulting in uniform mask modeling that cannot effectively represent the anatomical structure of the teeth, causing blurring of the boundary of the reconstructed images, further limiting the performance of this type of method in CBCT images. 

Motivated by the above, this work aims to maintain the superiority of sparse modeling of masked autoencoder with the introduction of tooth boundary prompt, which in turn facilitates the effective identification of tooth boundary features in the pre-training phase for high-performance semantic segmentation in the case of limited data. On a CBCT multi-class tooth segmentation dataset accurately labeled by experts, our self-supervised pre-training method outperforms other comparative methods, further demonstrating its viability.

Our main contributions can be summarized as:
\begin{enumerate}
    \item To introduce anatomical features in medical images in to MIM, we proposed a masked autoencoder-based prompt mechanism for embedding important anatomical features. 
    \item To maintain the feature sparsity advantage of the MAE paradigm, we designed a prompt branch using the Graph Attention Mechanism to effectively represent boundary features in a sparse manner.
    \item Extensive experiments have demonstrated that our method exhibits competitive results relative to current widely used self-supervised pre-training methods, further demonstrating its potential. 
   
\end{enumerate}
\begin{table*}[!t]
\label{table1}
\centering
	\fontsize{10}{4}\selectfont
	\begin{threeparttable}
		\caption{Comparison Results}
		\label{tab:perfor}
            \setlength{\tabcolsep}{3 mm}{
		\begin{tabular*}{\linewidth}{ccccccccc}
			\toprule
			Data$^*$&Methods&DSC&Jaccard&Precesion&Recall& HD (mm)\cr
			\midrule
			\multirow{5}*{Full}
			& UNetR (baseline) &$83.19\pm7.99$&$75.73\pm8.68$&$87.01\pm6.95$&$81.96\pm8.48$&$9.13\pm4.44$\cr
			~&
		       MAE~\cite{zhou2022self}
            &$87.32\pm5.13$&$81.29\pm5.97$&$92.27\pm3.87$&$85.41\pm5.86$&$2.92\pm1.67$\cr
			~&
			SimCLR~\cite{SimClR} &$87.83\pm5.20$&$81.83\pm6.11$&$91.30\pm2.86$&$86.95\pm6.13$&$2.92\pm0.80$\cr
			~&
			SimMIM~\cite{MIM} &$87.22\pm4.55$&$80.95\pm5.23$&$91.03\pm1.98$&$86.51\pm5.40$&$3.29\pm0.99$\cr
		    ~&
			\textbf{Ours}
			&\boldmath$89.78\pm2.14$&\boldmath$84.15\pm3.09$&\boldmath$93.88\pm0.93$&\boldmath$88.04\pm3.12$&\boldmath $2.13\pm0.87$\cr
                \midrule
			\multirow{5}*{Half}
			& UNetR (baseline) &$79.28\pm7.56$&$71.70\pm8.32$&$86.19\pm5.26$&$78.22\pm8.13$&$7.18\pm2.17$\cr
			~&
		       MAE
             &$82.15\pm4.99$&$74.35\pm6.04$&$86.63\pm3.04$&$81.32\pm5.86$&$6.72\pm2.21$\cr
			~&
			SimCLR &$82.57\pm6.75$&$75.27\pm7.62$&\boldmath$88.16\pm6.08$&$80.51\pm6.79$&\boldmath$6.11\pm1.95$\cr
			~&
			SimMIM &$81.69\pm6.62$&$74.13\pm7.32$&$86.20\pm5.20$&$80.31\pm7.37$&$6.94\pm2.97$\cr
		    ~&
                \textbf{Ours}
			&\boldmath$83.84\pm4.07$&\boldmath$75.76\pm4.82$&$86.95\pm3.08$&\boldmath$83.15\pm4.39$& $8.73\pm1.78$\cr
			\bottomrule

            \end{tabular*}
            }
\begin{tablenotes}
\item[*] We split the labeled data for training into 2 forms, full indicates 50 labeled training cases and half denotes 25 labeled training cases.
\end{tablenotes}
        \end{threeparttable}
\end{table*}

\section{METHOD}
\label{sec:format}
The schematic overview of this work is shown in Fig.~\ref{fig_frameworks}. The methodology in this work can be divided into 1) Graph attention-based sparse prompt branch, 2) Boundary prompt-based mask pre-training, and 3) Downstream segmentation.
\subsection{Graph Attention-based Sparse prompt branch}
\label{GA}

In this branch, we employ sparsely labeled tooth boundaries as training labels in a lightweight encoder-decoder paradigm. To efficiently capture valuable information from these sparse annotations while preserving their sparsity, we have designed the encoder architecture, drawing inspiration from the graph attention mechanism introduced by Li et al.~\cite{li2022semantic}. 

For an input batch $\textbf{X}^{\left(B\times W \times H \times D\right)}$, it will first be patched into $\textbf{X}_p^{\left(B\times N \times C\right)}$ reffered to~\cite{hatamizadeh2022unetr}. In accordance with the ISO tooth numbering criterion, we represent the relationships between teeth as an adjacency matrix same as~\cite{li2022semantic} and map $N$ to $\mathcal V=33$ to create the nodes of the graph attention network. Let $\mathcal{G} =\left (\mathcal V ,\mathcal E  \right ) $ be a graph where $\mathcal V $ is a collecting nodes for each teeth, $\mathcal E=\left \{A_{i,j}| v_i,v_j\in \mathcal{V}  \right \} $

Thus, the feed-forward inference operation for every node can be driven as follows:
\begin{align}
    \mathbf{x}^{\prime}_i = \alpha_{i,i}\mathbf{\Theta}_{s}\mathbf{x}_{i} +
        \sum_{j \in \mathcal{N}(i)}
        \alpha_{i,j}\mathbf{\Theta}_{t}\mathbf{x}_{j}
\end{align}
where the attention coefficient is calculated as follows,
\begin{align}
\alpha_{i,j} =
        \frac{
        \exp\left(\mathbf{a}^{\top}\sigma \left(
        \mathbf{\Theta}_{s} \mathbf{x}_i + \mathbf{\Theta}_{t} \mathbf{x}_j
        \right)\right)}
        {\sum_{k \in \mathcal{N}(i) \cup \{ i \}}
        \exp\left(\mathbf{a}^{\top}\sigma \left(
        \mathbf{\Theta}_{s} \mathbf{x}_i + \mathbf{\Theta}_{t} \mathbf{x}_k
        \right)\right)}
\end{align}
where $x_i$ and $x_{i}^{'}$ denotes an update of the node $v_i$ parameters. $\Theta$ represents the learnable parameter matrices. $N$ represents the set of all adjacent nodes of node $v_i$. $\sigma$ denotes the LeakyReLU activation function.
Therefore, The entire forward process in this phase can be represented as
\begin{align}
Output = Decoder(\sum_{i\in N} fc_2(GA_i(fc_1(x_p))+fc_1(x_p)))
\end{align}
Where $GA$ denotes the graph attention operation mentioned before. $Decoder$ is chosen to be the UNetR decoder. $fc_i$ denotes the full connection layer and $+$ represents the residual connection.

Since the positive samples at the tooth boundary are very sparse and unbalanced with respect to the background, here, we choose the Tversky loss~\cite{loss1} as the loss function.

\subsection{Boundary Prompt-based Mask Pre-training}
\label{pretrain}
To incorporate anatomical tooth boundary information into the MAE training process without introducing extra training parameters, we adopt a strategy where we freeze the parameters of the prompt branch and leverage its output as masked tokens. By contrasting these refined masked tokens with original images, our network enhances its ability to capture crucial anatomical features of teeth, thereby facilitating the achievement of task-specific self-supervised pre-training goals. Specifically, we first perform the same patchization of the input image as in Section~\ref{GA}, Patches are simultaneously input into both the encoder of UNetR and the prompt branch, resulting in the generation of the original encoding and the mask. The extent to which the original encoding is replaced by the mask is determined by the mask rate, after which the decoding process takes place. The primary objective at this stage is to reconstruct and restore the mask to its original image form.

In detail, for input patch $\textbf{X}_p^{\left(B\times N \times C\right)}$,
it will be fed into the UNetR encoder to generate original tokens $X_t$ and mask tokens $X_{mask}$. Thus, our mask strategy can be represented as,
\begin{align}
\hat{x}_p = \sigma(Decoder (x_p \odot x_{\text{mask} }\cdot \alpha ))\label{eq:mask}
\end{align}
where $\hat{x_p}$ represents the reconstructed output and $\alpha$ represents the mask rate.

The network task in this phase is to reconstruct the mask image to the original image in a self-supervised manner, which can be represented as,
\begin{align}
    \min_{\boldsymbol{\theta}} \text{MSE}(\boldsymbol{\theta}) = \frac{1}{N} \sum_{i=1}^{N} (y_i - f(\hat{x}_p, \boldsymbol{\theta}))^2 
\end{align}
Where $y_i$ denotes the original input image.

\subsection{Downstream Segmentation}

During this stage, we initialize the model with the pre-training weights acquired in Section~\ref{pretrain}. We then employ a limited set of labeled data for supervised training. Given the requirement in multi-class semantic segmentation to account for both background and foreground, as well as distinctions among various foreground categories, we formulate the loss function as follows,
\begin{align}
\mathcal{L}_{seg} = -\left( \beta \cdot \text{BCE}(y, \hat{y}) + (1 - \beta) \cdot \text{Dice}(y, \hat{y}) \right) \label{eq:loss}
\end{align}
Where BCE and Dice mean Cross entropy loss and dice loss.
\begin{figure*}[!t]
\centering
\includegraphics[width=0.95\textwidth]{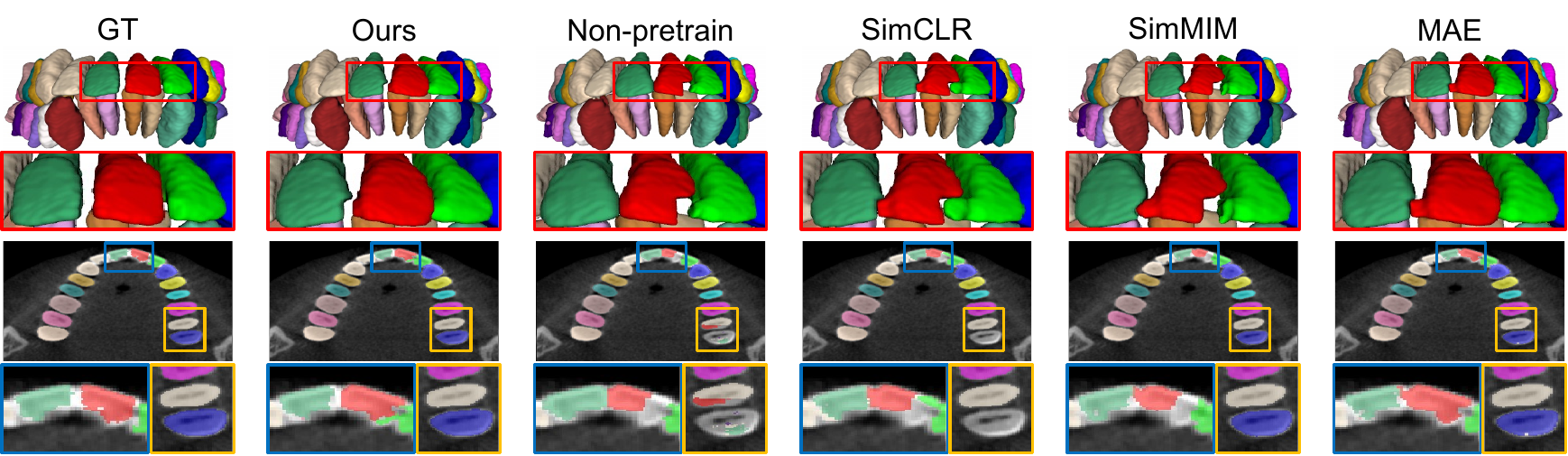}
\caption{Comparing results of the 2D and 3D visualizations. The results demonstrate that in the case of incisors in close proximity and wisdom teeth, our method effectively avoids boundary blurred and missed segmentation.}
\label{fig_framework}
\end{figure*}
\section{EXPERIMENTS}
\label{sec:pagestyle}

\subsection{Experiment setting}
The training and testing phases of the proposed pretraining and segmentation network were executed end-to-end on a single NVIDIA RTX A6000 GPU using PyTorch. We configured the batch size and iteration step to 2 and 10,000, respectively, to facilitate gradient updates. The initial learning rate was set to 1e-4, and we applied a learning rate decay of 0.1 every 2,500 steps. We employed the Adam optimizer to minimize the loss function during segmentation.

\subsection{Data and Evaluation Metrics}
We assess the efficacy of our method using a dataset sourced from dental clinics. 
This dataset includes cases involving missing teeth, crowding, oblique positions, and metal artifacts. 
Comprising 158 clinical dental CBCT scans, the dataset was meticulously curated by accomplished medical professionals at the Stomatological Hospital of Chongqing Medical University, complete with manually annotated labels. 
We partitioned the dataset into four distinct sets: a pre-training set, a training set, a validation set, and a testing set. For the initial boundary pre-training phase, we employed 100 cases with sparse boundary labeling. Subsequently, we utilized 50/25 out of the 100 cases with full annotations for the training set. To assess our model's performance, 28 cases were designated for the validation set, while the final 30 cases constituted the test set.

The voxel resolution of the dental CBCT image data ranged from 0.25 mm × 0.25 mm × 0.25 mm to 0.4 mm × 0.4 mm × 0.4 mm, with each slice measuring 400 × 400 pixels. The number of slices varied from 124 to 328, and the intensity values fell within the range of -1000 to 8000, albeit unevenly distributed.

In this study, we adopt the 5 standard medical segmentation metrics Dice Score Coefficient (DSC), the 95th percentile Hausdorff Distance (HD95), Jaccard Distance, Precision and Recall to evaluate our proposed semi-supervised approach.

\subsection{Results} 
For our proposed methodology, we have selected UNetR as the foundational semantic segmentation architecture due to its outstanding performance and simplicity. Specifically, we utilized the Vision Transformer as the backbone network for comparative analysis. In our semi-supervised experiments, we employed native CBCT images as unlabeled data. We conducted a comparative study with four other algorithms based on these backbone networks, including a supervised algorithm and three semi-supervised algorithms that do not take into account the specific morphological characteristics of teeth.

In the experiments, we applied different data splits for labeled images, including full and half partitions. As illustrated in Table~\ref{table1}, under both data volume scenarios, our algorithm yielded the best Dice scores, surpassing the quantitative metrics of the supervised method and other semi-supervised methods by approximately 2\%. Furthermore, to underscore the clinical relevance of our approach, visual examples of all algorithms are provided in Figure~\ref{fig_framework}.
As demonstrated in the figure, the regions delineated by red and blue rectangles and delineated by yellow rectangle represent cases involving clinically incisors in close proximity and wisdom teeth scenarios, respectively, often entailing boundary blurring and classification confusion. In the visual results, our method adeptly solved these challenging scenarios.

\section{CONCLUSION}
\label{sec:majhead}

In this work, we propose a novel Mask Reconstruction Pre-trained mechanism and Graph Attention-Based Tooth Boundary Prompt Branch to enhance the segmentation performance and generalizability of CBCT tooth segmentation models under limited data conditions. Extensive experiment shows that our method perform competitive results in training scenarios with different data volumes.
For future work, we hope to further optimize the framework to enable end-to-end deployment for clinical use. At the same time, we would like to introduce more essential dental anatomical features for pre-training rather than tooth boundaries.

\section{Compliance with ethical standards}
\label{sec:ethics}
All human volunteer study protocols followed the Institutional Review Board of Stomatological Hospital of Chongqing Medical University's ethical requirements, Helsinki Declaration. Before conducting this work, the Stomatological Hospital of Chongqing Medical University Biomedical Institutional Review Board gave ethical approval. This paper did not require written informed permission because all patients were retrospectively gathered.

\section{Acknowledgments}
\label{sec:acknowledgments}
This work was supported in part by the National Natural Science Foundation of China [grant numbers 62206036, 82101058]; the Natural Science Foundation of Chongqing, China [grant number cstc2020jcyj-msxmX0525].
\bibliographystyle{IEEEbib}
\bibliography{refs}

\end{document}